\documentclass[final,authoryear,5p,times]{elsarticle}

\usepackage{epsfig}
\usepackage{graphicx}
\usepackage{amsmath}
\usepackage{amssymb}
\usepackage{verbatim}
\usepackage{algorithm}
\usepackage{algorithmic}
\usepackage{booktabs}
\usepackage{multirow}
\usepackage{pbox} 
\usepackage{tabularx}
\usepackage{changepage}                 
\usepackage{lipsum}
\usepackage{tabularx}
\usepackage{hyperref}       
\usepackage{url}            
\usepackage{bm} 
\usepackage{color}
\usepackage{subcaption}
\newlength{\offsetpage}
	{\end{adjustwidth}}

\journal{Elsevier}

\begin{document}
\begin{frontmatter}



\title{Aleatoric uncertainty estimation with test-time augmentation for medical image segmentation with convolutional neural networks  }



\author{Guotai Wang\corref{cor1}$^{a,b,c}$}
\cortext[cor1]{Corresponding author}
\ead{guotai.1.wang@kcl.ac.uk}
\author{Wenqi Li$^{a,b}$, Michael Aertsen$^{d}$, Jan Deprest$^{a,d,e,f}$, S\'ebastien~Ourselin$^b$, Tom Vercauteren$^{a,b,f}$}

\address{
	$^a$Wellcome / EPSRC Centre for Interventional and Surgical Sciences, University College London, London, UK\\
	$^b$School of Biomedical Engineering and Imaging Sciences, King's College London, London, UK\\
	$^c$School of Mechanical and Electrical Engineering, University of Electronic Science and Technology of China, Chengdu, China\\
	$^d$Department of Radiology, University Hospitals Leuven, Leuven, Belgium\\
	$^e$Institute for Women's Health, University College London, London, UK \\
	$^f$Department of Obstetrics and Gynaecology, University Hospitals Leuven, Leuven, Belgium \\

}
\begin{abstract}
 Despite the state-of-the-art performance for medical image segmentation, deep convolutional neural networks (CNNs)  have rarely provided uncertainty estimations regarding their segmentation outputs, e.g.,  model (\textit{epistemic}) and image-based (\textit{aleatoric}) uncertainties. In this work, we analyze these different types of uncertainties for CNN-based 2D and 3D medical image segmentation tasks at both pixel level and structure level. We additionally propose a test-time augmentation-based \textit{aleatoric} uncertainty to analyze the effect of different transformations of the input image on the segmentation output. Test-time augmentation has been previously used to improve segmentation accuracy, yet not been formulated in a consistent mathematical framework. Hence, we also propose a theoretical formulation of test-time augmentation, where a distribution of the prediction is estimated by Monte Carlo simulation with prior distributions of parameters in an image acquisition model that involves image transformations and noise. We compare and combine our proposed \textit{aleatoric} uncertainty with model uncertainty. Experiments with segmentation of fetal brains and brain tumors from 2D and 3D Magnetic Resonance Images (MRI) showed that 1) the test-time augmentation-based \textit{aleatoric} uncertainty provides a better uncertainty estimation than calculating the test-time dropout-based model uncertainty alone and helps to reduce overconfident incorrect predictions, and 2) our test-time augmentation outperforms a single-prediction baseline and dropout-based multiple predictions.    

\end{abstract}


\end{frontmatter}


\section{Introduction}
Segmentation of medical images is an essential task for many applications such as anatomical structure modeling, tumor growth measurement, surgical planing and treatment assessment~\citep{Sharma2010}. Despite the breadth and depth of current research, it is very challenging to achieve accurate and reliable segmentation results for many targets~\citep{Withey2007}. This is often due to poor image quality, inhomogeneous appearances brought by pathology, various imaging protocols and large variations of the segmentation target among patients. Therefore, uncertainty estimation of segmentation results is critical for understanding how reliable the segmentations are. For example, for many images, the segmentation results of pixels near the boundary are likely to be uncertain because of the low contrast between the segmentation target and surrounding tissues, where uncertainty information of the segmentation can be used to indicate potential mis-segmented regions or guide user interactions for refinement~\citep{Prassni2010,Wang2018}.   

In recent years, deep learning with convolutional neural networks (CNN) has achieved the state-of-the-art performance for many medical image segmentation tasks~\citep{Milletari2016, Abdulkadir2016, Kamnitsas2017}. Despite their impressive performance and the ability of automatic feature learning, these approaches do not by default provide uncertainty estimation for their segmentation results. In addition, having access to a large training set plays an important role for deep CNNs to achieve human-level performance ~\citep{Esteva2017,Rajpurkar2017}. However, for medical image segmentation tasks, collecting a very large dataset with pixel-wise annotations for training is usually difficult and time-consuming. As a result, current medical image segmentation methods based on deep CNNs use relatively small datasets compared with those for natural image recognition\citep{Russakovsky2015a}. This is likely to introduce more uncertain predictions for the segmentation results, and also leads to uncertainty of downstream analysis, such as volumetric measurement of the target. Therefore, uncertainty estimation is highly desired for deep CNN-based medical image segmentation methods. 

Several works have investigated uncertainty estimation for deep neural networks~\citep{Kendall2017, Lakshminarayanan2017,Zhu2018,Ayhan2018}. They focused mainly on image classification or regression  tasks, where the prediction outputs are high-level image labels or bounding box parameters, therefore uncertainty estimation is usually only given for the high-level predictions. In contrast, pixel-wise predictions are involved in segmentation tasks, therefore pixel-wise uncertainty estimation is highly desirable. In addition, in most interactive segmentation cases, pixel-wise uncertainty information is more helpful for intelligently guiding the user to give interactions. However, previous works have rarely demonstrated uncertainty estimation for deep CNN-based medical image segmentation. As suggested by ~\cite{Kendall2017}, there are two major types of predictive uncertainties for deep CNNs: \textit{epistemic} uncertainty and \textit{aleatoric} uncertainty. \textit{Epistemic} uncertainty is also known as model uncertainty that can be explained away given enough training data, while \textit{aleatoric} uncertainty depends on noise or randomness in the input testing image. 

In contrast to previous works focusing mainly on classification or regression-related uncertainty estimation, and recent works of ~\cite{Nair2018} and~\cite{Roy2018a} investigating only test-time dropout-based (\textit{epistemic}) uncertainty for segmentation, we extensively investigate different kinds of uncertainties for CNN-based medical image segmentation, including not only \textit{epistemic} but also \textit{aleatoric} uncertainties for this task. We also propose a more general estimation of \textit{aleatoric} uncertainty that is related to not only  image noise but also spatial transformations of the input, considering different possible poses of the object during image acquisition. To obtain the transformation-related uncertainty, we augment the input image at test time, and obtain an estimation of the distribution of the prediction based on test-time augmentation. Test-time augmentation (e.g., rotation, scaling, flipping) has been recently used to improve performance of image classification~\citep{Matsunaga2017} and nodule detection~\citep{Jin2018}. ~\cite{Ayhan2018} also showed its utility for uncertainty estimation in a fundus image classification task. However, the previous works have not provided a mathematical or theoretical formulation for this. Motivated by these observations, we propose a mathematical formulation for test-time augmentation, and analyze its performance for the general ~\textit{aleatoric} uncertainty estimation in medical image segmentation tasks. In the proposed formulation, we represent an image as a result of an acquisition process which involves geometric transformations and image noise. We model the hidden parameters of the image acquisition process with prior distributions, and predict the distribution of the output segmentation for a test image with a Monte Carlo sampling process. With the samples from the distribution of the predictive output based on the same pre-trained CNN, the variance/entropy can be calculated for these samples, which provides an estimation of the \textit{aleatoric} uncertainty for the segmentation.

The contribution of this work is three-fold. First, we propose a theoretical formulation of test-time augmentation for deep learning. Test-time augmentation has not been mathematically formulated by existing works, and our proposed mathematical formulation is general for image recognition tasks. Second, with the proposed formulation of test-time augmentation, we propose a general \textit{aleatoric} uncertainty estimation for medical image segmentation, where the uncertainty comes from not only image noise but also spatial transformations. Third, we analyze different types of uncertainty estimation for the deep CNN-based segmentation, and validate the superiority of the proposed general \textit{aleatoric} uncertainty with both 2D and 3D segmentation tasks. 

\section{Related Works}
\subsection{Segmentation Uncertainty}
Uncertainty estimation has been widely investigated for many existing medical image segmentation tasks. As way of examples, \cite{Saad2010} used shape and appearance prior information to estimate the uncertainty for probabilistic segmentation of medical imaging. \cite{Shi2012} estimated the uncertainty of graph cut-based cardiac image segmentation, which was used to improve the robustness of the system. \cite{Sankaran2015} estimated lumen segmentation uncertainty for realistic patient-specific blood flow modeling.  \cite{Parisot2014} used segmentation uncertainty to guide content-driven adaptive sampling for concurrent brain tumor segmentation and registration. \cite{Prassni2010} visualized the uncertainty of a random walker-based segmentation to guide volume segmentation of brain Magnetic Resonance Images (MRI) and abdominal Computed Tomography (CT) images. \cite{Top2011} combined uncertainty estimation with active learning to reduce user time for interactive 3D image segmentation. 

\subsection{Uncertainty Estimation for Deep CNNs}
For deep CNNs, both \textit{epistemic} and \textit{aleatoric} uncertainties have been investigated in recent years. For model (\textit{epistemic}) uncertainty, exact Bayesian networks offer a mathematically grounded method, but they are hard to implement and computationally expensive. Alternatively, it has been shown that dropout at test time can be cast as a Bayesian approximation to represent model uncertainty~\citep{Gal2016, Li2017}. \cite{Zhu2018} used Stochastic Variational Gradient Descent (SVGD) to perform approximate Bayesian inference on uncertain CNN parameters. A variety of other approximation methods such as Markov chain Monte Carlo (MCMC)~\citep{Neal2012}, Monte Carlo Batch Normalization (MCBN) ~\citep{Teye2018} and variational Bayesian methods \citep{Graves2011,Louizos2016} have also been developed. \cite{Lakshminarayanan2017} proposed ensembles of multiple models for uncertainty estimation, which was simple and scalable to implement. For test image-based (\textit{aleatoric}) uncertainty, \cite{Kendall2017} proposed a unified Bayesian deep learning framework to learn mappings from input data to \textit{aleatoric} uncertainty and composed them with \textit{epistemic} uncertainty, where the \textit{aleatoric} uncertainty was modeled as learned loss attenuation and further categorized into \textit{homoscedastic} uncertainty and \textit{heteroscedastic} uncertainty. \cite{Ayhan2018} used test-time augmentation for \textit{aleatoric} uncertainty estimation, which was an efficient and effective way to explore the locality of a testing sample.  However, its utility for medical image segmentation has not been demonstrated.

\subsection{Test-Time Augmentation}
Data augmentation was originally proposed for the training of deep neural networks. It was employed to enlarge a relatively small dataset by applying transformations to its samples to create new ones for training~\citep{Krizhevsky2012}. The transformations for augmentation typically include flipping, cropping, rotating, and scaling training images. \cite{Abdulkadir2016} and \cite{Hefny2015a} also used elastic deformations for biomedical image segmentation. Several studies have empirically found that combining predictions of multiple transformed versions of a test image helps to improve the performance. For example, \cite{Matsunaga2017}  geometrically transformed test images for skin lesion classification. \cite{Radosavovic2017} used a single model to predict multiple transformed copies of unlabeled images for data distillation. \cite{Jin2018} tested on samples extended by rotation and translation for pulmonary nodule detection. However, all these works used test-time augmentation as an ad hoc method, without detailed formulation or theoretical explanation, and did not apply it to uncertainty estimation for segmentation tasks.   

\section{Method}
The proposed general \textit{aleatoric} uncertainty estimation is formulated in a consistent mathematical framework including two parts. The first part is a mathematical representation of ensembles of predictions of multiple transformed versions of the input. We represent an image as a result of an image acquisition model with hidden parameters in Section~\ref{image_acq_model}. Then we formulate test-time augmentation as inference with hidden parameters following given prior distributions in Section~\ref{inference_aug}. The second part calculates the diversity of the prediction results of an augmented test image, and it is used to estimate the \textit{aleatoric} uncertainty related to image transformations and noise. This is detailed in Section~\ref{data_based_uncertainty}. Our proposed \textit{aleatoric} uncertainty is compared and combined with \textit{epistemic} uncertainty, which is described in Section~\ref{epistemic_uncertainty}.
Finally, we apply our proposed method to structure-wise uncertainty estimation in Section~\ref{sec:strcture-wise}.

\subsection{Image Acquisition Model} \label{image_acq_model} 
The image acquisition model describes the process by which the observed images have been obtained. This process is confronted with a lot of factors that can be related or unrelated to the imaged object, such as blurring, down-sampling, spatial transformation, and system noise. While blurring and down-sampling are commonly considered for image super-resolution \citep{Yue2016}, in the context of image recognition they have a relatively lower impact. Therefore, we focus on the spatial transformation and noise, and highlight that adding more complex intensity changes or other forms of data augmentation such as elastic deformations is a straightforward extension. The image acquisition model is:
\begin{align}\label{eq:image_model}
X = \mathcal{T}_{\bm\beta}(X_0) + \bm e
\end{align}
where $X_0$ is an underlying image in a certain position and orientation, i.e., a hidden variable. $\mathcal{T}$ is a transformation operator that is applied to $X_0$. $\bm \beta$ is the set of parameters of the transformation, and $\bm e$ represents the noise that is added to the transformed image. $X$ denotes the observed image that is used for inference at test time. Though the transformations can be in spatial, intensity or feature space, in this work we only study the impact of reversible spatial transformations (e.g., flipping, scaling, rotation and translation), which are the most common types of transformations occurring during image acquisition and used for data augmentation purposes. Let $\mathcal{T}^{-1}_{\bm\beta}$ denote the inverse transformation of $\mathcal{T}_{\bm\beta}$, then we have:

\begin{align}
\label{eq:image_model_reverse}
X_0 = \mathcal{T}^{-1}_{\bm\beta}(X -\bm e)
\end{align}  
Similarly to data augmentation at training time, we assume that the distribution of $X$ covers the distribution of $X_0$. In a given application, this assumption leads to some prior distributions of the transformation parameters and noise. For example, in a 2D slice of fetal brain MRI, the orientation of the fetal brain can span all possible directions in a 2D plane, therefore the rotation angle $\bm r$ can be modeled with a uniform prior distribution $\bm r \sim U(0, 2\pi)$. The image noise is commonly modeled as a Gaussian distribution, i.e., $\bm e \sim \mathcal{N}(\bm \mu, \bm \sigma)$, where $\bm \mu$ and $\bm \sigma$ are the mean and standard deviation respectively. Let $p({\bm\beta})$ and $p({\bm e})$ represent the prior distribution of $\bm \beta$ and $\bm e$ respectively, therefore we have $\bm \beta \sim p({\bm\beta})$ and $\bm e \sim p({\bm e})$. 

Let $Y$ and $Y_0$ be the labels related to $X$ and $X_0$ respectively. For image classification, $Y$ and $Y_0$ are categorical variables, and they should be invariant with regard to transformations and noise, therefore $Y = Y_0$. For image segmentation, $Y$ and $Y_0$ are discretized label maps, and they are equivariant with the spatial transformation, i.e., $Y = \mathcal{T}_{\bm\beta}(Y_0)$.
\subsection{Inference with Hidden Variables}\label{inference_aug}
In the context of deep learning, let $f(\cdot)$ be the function represented by a neural network, and $\bm \theta$ represent the parameters learned from a set of training images with their corresponding annotations. In a standard formulation, the prediction $Y$ of a test image $X$ is inferred by:
\begin{align}\label{eq:network}
Y = f(\bm \theta, X)
\end{align}    
For regression problems, $Y$ refers to continuous values. For segmentation or classification problems, $Y$ refers to discretized labels obtained by \textit{argmax} operation in the last layer of the network.  Since $X$ is only one of many possible observations of the underlying image $X_0$, direct inference with $X$ may lead to a biased result affected by the specific transformation and noise associated with $X$. To address this problem, we aim at inferring it with the help of the latent $X_0$ instead:
\begin{align}\label{eq:inference_hidden}
Y = \mathcal{T}_{\bm\beta}(Y_0) = \mathcal{T}_{\bm\beta}\big(f\big(\bm \theta, X_0)\big) =  \mathcal{T}_{\bm\beta}\Big(f\big(\bm \theta, \mathcal{T}^{-1}_{\bm\beta}(X -\bm e)\big)\Big)
\end{align}   
where the exact values of $\bm \beta$ and $\bm e$ for $X$ are unknown. Instead of finding a deterministic prediction of $X$, we alternatively consider the distribution of $Y$ for a robust inference given the distributions of $\bm \beta$ and $\bm e$.
\begin{align}\label{eq:prob_y}
p(Y|X) = p\Bigg(\mathcal{T}_{\bm\beta}\Big(f\big(\bm \theta, \mathcal{T}^{-1}_{\bm\beta}(X -\bm e)\big)\Big)\Bigg), \text{where } \bm \beta \sim p({\bm \beta}), \bm e \sim p({\bm e})
\end{align}
For regression problems, we obtain the final prediction for $X$ by calculating the expectation of $Y$ using the distribution $p(Y|X)$.
\begin{equation}
\begin{aligned}\label{eq:expectation_y}
E(Y|X) & = \int y p(y|X) dy \\
& = \int_{\bm \beta \sim p({\bm \beta}), \bm e \sim p({\bm e})} \mathcal{T}_{\bm\beta}\Big(f\big(\bm \theta, \mathcal{T}^{-1}_{\bm\beta}(X -\bm e)\big)\Big)p(\bm \beta)p(\bm e) d\bm \beta d \bm e
\end{aligned}
\end{equation}
Calculating $E(Y|X)$ with Eq. \eqref{eq:expectation_y} is computationally expensive, as $\bm \beta$ and $\bm e $ may take continuous values and $p({\bm \beta})$ is a complex joint distribution of different types of transformations. Alternatively, we estimate $E(Y|X)$ by using Monte Carlo simulation. Let $N$ represent the total number of simulation runs. In the $n$-th simulation run, the prediction is:
\begin{equation}
y_n = \mathcal{T}_{\bm\beta_n}\Big(f\big(\bm \theta, \mathcal{T}^{-1}_{\bm\beta_n}(X -\bm e_n)\big)\Big)
\end{equation}
where $\bm \beta_n\sim p({\bm \beta}),~\bm e_n \sim p({\bm e})$. To obtain $y_n$, we first randomly sample $\bm \beta_n$ and $\bm e_n$ from the prior distributions $p({\bm \beta})$ and $p({\bm e})$, respectively. Then we obtain one possible hidden image with $\bm \beta_n$ and $\bm e_n$ based on Eq.~\eqref{eq:image_model_reverse}, and feed it into the trained network to get its prediction, which is transformed with $\bm \beta_n$ to obtain $y_{n}$  according to Eq.~\eqref{eq:inference_hidden}. With the set $\mathcal{Y} = \{y_1, y_2, ..., y_N\}$ sampled from $p(Y|X)$, $E(Y|X)$ is estimated as the average of $\mathcal{Y}$ and we use it as the final prediction $\hat{Y}$ for $X$:
\begin{equation}
\label{eq:average_y}
\hat{Y} = E(Y|X) \approx \frac{1}{N}\sum_{n=1}^N y_n 
\end{equation}
For classification or segmentation problems, $p(Y|X)$ is a discretized distribution. We obtain the final prediction for $X$ by maximum likelihood estimation:
\begin{equation}
\label{eq:maximual_likelihood_y}
\hat{Y} = \underset{y}{\arg \max}~p(y|X) \approx \text{Mode} \Big(\mathcal{Y}\Big)
\end{equation}
where Mode($\mathcal{Y}$) is the most frequent element in $\mathcal{Y}$. This corresponds to majority voting of multiple predictions. 

\subsection{Aleatoric Uncertainty Estimation with Test-Time Augmentation} \label{data_based_uncertainty}
The uncertainty is estimated by measuring how diverse the predictions for a given image are. Both the variance and entropy of the distribution $p(Y|X)$ can be used to estimate uncertainty. However, variance is not sufficiently representative in the context of multi-modal distributions. In this paper we use entropy for uncertainty estimation:
\begin{align}\label{eq:entropy}
H(Y|X) = - \int p(y|X)\text{ln}\big(p(y|X)\big) dy
\end{align} 
With the Monte Carlo simulation in Section~\ref{inference_aug}, we can approximate $H(Y|X)$ from the simulation results $\mathcal{Y} = \{y_1, y_2, ..., y_N\}$. Suppose there are $M$ unique values in $\mathcal{Y}$. For classification tasks, this typically refers to $M$ labels. Assume the frequency of the $m$-th unique value is $\hat p_m$, then $H(Y|X)$ is approximated as:
\begin{align}\label{eq:entropy_approx}
H(Y|X) \approx - \sum_{m=1}^{M} \hat p_m \text{ln}(\hat p_m)
\end{align} 
For segmentation tasks, pixel-wise uncertainty estimation is desirable. Let $Y^i$ denote the predicted label for the $i$-th pixel. With the Monte Carlo simulation, a set of values for $Y^i$ are obtained $\mathcal{Y}^i = \{y^i_1, y^i_2, ..., y^i_N\}$. The entropy of the distribution of $Y^i$ is therefore approximated as:
\begin{align}\label{eq:entropy_approx_seg}
H(Y^i|X) \approx - \sum_{m=1}^{M} \hat p^i_m \text{ln} (\hat p^i_m)
\end{align} 
where $\hat p^i_m$ is the frequency of the $m$-th unique value in $\mathcal{Y}^i$.

\subsection{Epistemic Uncertainty Estimation} \label{epistemic_uncertainty}
To obtain model (\textit{epistemic}) uncertainty estimation, we follow the test-time dropout method proposed by \cite{Gal2016}. In this method, let $q(\bm{\theta})$ be an approximating distribution over the set of network parameters $\bm \theta$ with its elements randomly set to zero according to Bernoulli random variables. $q(\bm{\theta})$ can be achieved by minimizing the Kullback-Leibler divergence between $q(\bm{\theta})$ and the posterior distribution of $\bm{\theta}$ given a training set. After training, the predictive distribution of a test image $X$ can be expressed as:
\begin{align}\label{eq:prob_y_epistemic}
p(Y|X) = \int p(Y|X,\omega)q(\omega) d\omega
\end{align}
The distribution of the prediction can be sampled based on Monte Carlo samples of the trained network (i.e, MC dropout): 
$y_n = f(\bm \theta_n, X) $ where $\bm \theta_n$ is a Monte Carlo sample from $q(\bm \theta)$. Assume the number of samples is $N$, and the sampled set of the distribution of $Y$ is $\mathcal{Y} = \{y_1, y_2, ..., y_N\}$. The final prediction for $X$ can be estimated by Eq.~\eqref{eq:average_y} for regression problems or Eq.~\eqref{eq:maximual_likelihood_y} for classification/segmentation problems. The \textit{epistemic} uncertainty estimation can therefore be calculated based on variance or entropy of the sampled $N$ predictions. To keep consistent with our \textit{aleatoric} uncertainty, we use entropy for this purpose, which is similar to Eq.~\eqref{eq:entropy_approx_seg}.  Test-time dropout may be interpreted as a way of ensembles of networks for testing. In the work of \cite{Lakshminarayanan2017}, ensembles of neural networks was explicitly proposed as an alternative solution of test-time dropout for estimating \textit{epistemic} uncertainty.

\subsection{Structure-wise Uncertainty Estimation}\label{sec:strcture-wise}
\cite{Nair2018} and \cite{Roy2018a} used Monte Carlo samples generated by test-time dropout for structure/lesion-wise uncertainty estimation. Following these works, we extend the structure-wise uncertainty estimation method by using Monte Carlo samples generated by not only test-time dropout, but also test-time augmentation described in ~\ref{inference_aug}. For $N$ samples from the Monte Carlo simulation, let $\mathcal{V} = \{v_1, v_2, ..., v_N\}$ denote the set of volumes of the segmented structure, where $v_i$ is the volume of the segmented structure in the $i$-th simulation. Let $\mu_{\mathcal{V}}$ and $\sigma_{\mathcal{V}}$ denote the mean value and standard deviation of $\mathcal{V}$ respectively.  We use the volume variation coefficient (VVC) to estimate the structure-wise uncertainty:
\begin{align}\label{eq:vvc}
VVC= \frac{\sigma_{\mathcal{V}}}{\mu_{\mathcal{V}}}
\end{align}
where VVC is agnostic to the size of the segmented structure.

\section{Experiments and Results}

\label{headings}

We validated our proposed testing and uncertainty estimation method with two segmentation tasks: 2D fetal brain segmentation from MRI slices and 3D brain tumor segmentation from multi-modal MRI volumes. The implementation details for 2D and 3D segmentation are described in Section~\ref{2D_exepriment} and Section~\ref{3D_exepriment} respectively.

In both tasks, we compared different types of uncertainties for the segmentation results: 1) the proposed \textit{aleatoric} uncertainty based on our formulated test-time augmentation (TTA), 2) the \textit{epistemic} uncertainty based on test-time dropout (TTD) described in Section~\ref{epistemic_uncertainty}, and 3) hybrid uncertainty that combines the \textit{aleatoric} and \textit{epistemic} uncertainties based on TTA + TTD. For each of these three methods, the uncertainty was obtained by Eq.~\eqref{eq:entropy_approx_seg} with $N$ predictions. For TTD and TTA + TTD, the dropout probability was set as a typical value of 0.5~\citep{Gal2016}.

We also evaluated the segmentation accuracy of these different prediction methods: TTA, TTD, TTA + TTD and the baseline that uses a single prediction without TTA and TTD. For a given training set, all these methods used the same model that was trained with data augmentation and dropout at training time. The augmentation during training followed the same formulation in Section \ref{image_acq_model}. We investigated the relationship between each type of uncertainty and segmentation error in order to know which uncertainty has a better ability to indicate potential mis-segmentations. Quantitative evaluations of segmentation accuracy are based on Dice score and Average Symmetric Surface Distance (ASSD). 
\begin{align}\label{eq:dice}
	Dice = \frac{2\times TP}{
	2\times TP + FN + FP}
\end{align}
where $TP$, $FP$ and $FN$ are true positive, false positive and false negative respectively. The definition of ASSD is:
\begin{align}\label{eq:assd}
ASSD = \frac{1}{|S| + |G|}\bigg(\sum_{s\in S}d(s,G) + \sum_{g\in G}d(g,S)\bigg)
\end{align}
where $S$ and $G$ denote the set of surface points of a segmentation result and the ground truth respectively. $d(s,G)$ is the shortest Euclidean distance between a point $s\in S$ and all the points in $G$. 
\subsection{2D Fetal Brain Segmentation from MRI}

\label{2D_exepriment}
Fetal MRI has been increasingly used for study of the developing fetus as it provides a better soft tissue contrast than the widely used prenatal sonography. The most commonly used imaging protocol for fetal MRI is Single-Shot Fast Spin Echo (SSFSE) that acquires images in a fast speed and mitigates the effect of fetal motion, leading to stacks of thick 2D slices. Segmentation is a fundamental step for fetal brain study, e.g., it plays an important role in inter-slice motion correction and high-resolution volume reconstruction~\citep{Tourbier2017, Ebner2018a}. Recently, CNNs have achieved the state-of-the-art performance for 2D fetal brain segmentation~\citep{Rajchl2016a, Salehi2017,Salehi2017isbi}. In this experiment, we segment the 2D fetal brain using deep CNNs with uncertainty estimation.
\subsubsection{Data and Implementation}

We collected clinical T2-weighted MRI scans of 60 fetuses in the second trimester with SSFSE on a 1.5 Tesla MR system (Aera, Siemens, Erlangen, Germany). The data for each fetus contained three stacks of 2D slices acquired in axial, sagittal and coronal views respectively, with pixel size 0.63 mm to 1.58 mm and slice thickness 3 mm to 6 mm. The gestational age ranged from 19 weeks to 33 weeks. We used 2640 slices from 120 stacks of 40 patients for training, 278 slices from 12 stacks of 4 patients for validation and 1180 slices from 48 stacks of 16 patients for testing. 
Two radiologists manually segmented the brain region for all the stacks slice-by-slice, where one radiologist gave a segmentation first, and then the second senior radiologist refined the segmentation if disagreement existed, the output of which were used as the ground truth. We used this dataset for two reasons. First, our dataset fits with a typical medical image segmentation application where the number of annotated images is limited. This leads the uncertainty information to be of high interest for robust prediction and our downstream tasks such as fetal brain reconstruction and volume measurement. Second, the position and orientation of fetal brain have large variations, which is suitable for investigating the effect of data augmentation.
For preprocessing, we normalized each stack by its intensity mean and standard deviation, and resampled each slice with pixel size 1.0 mm.

We used 2D networks of Fully Convolutional Network (FCN) \citep{Long2014}, U-Net\citep{Hefny2015a} and P-Net\citep{Wang2018}. The networks were implemented in TensorFlow\footnote{\url{https://www.tensorflow.org}}\citep{Abadi2016} using NiftyNet\footnote{\url{http://www.niftynet.io}}\citep{Li2017, Gibson2018}. During training, we used Adaptive Moment Estimation (Adam) to adjust the learning rate that was initialized as $10^{-3}$, with batch size 5, weight decay $10^{-7}$ and iteration number $10k$. We represented the transformation parameter $\bm\beta$ in the proposed augmentation framework as a combination of $f_l$, $r$ and $s$, where $f_l$ is a random variable for flipping along each 2D axis, $r$ is the rotation angle in 2D, and $s$ is a scaling factor. The prior distributions of these transformation parameters and random intensity noise were modeled as $f_l \sim Bern(\mu_f)$, $r\sim U(r_0, r_1)$, $s\sim U(s_0, s_1)$ and $ e \sim N(\mu_e, \sigma_e)$. The hyper-parameters for our fetal brain segmentation task were set as $\mu_f = 0.5$, $r_0 = 0$, $r_1 = 2\pi$, $s_0 = 0.8$ and $s_1 = 1.2$. For the random noise, we set $\mu_e = 0$ and $\sigma_e = 0.05$, as a median-filter smoothed version of a normalized image in our dataset has a standard deviation around 0.95.
We augmented the training data with this formulation, and during test time, TTA used the same prior distributions of augmentation parameters as used for training. 

\begin{figure*}[t]
	\centering
	\includegraphics[width=1.0\linewidth]{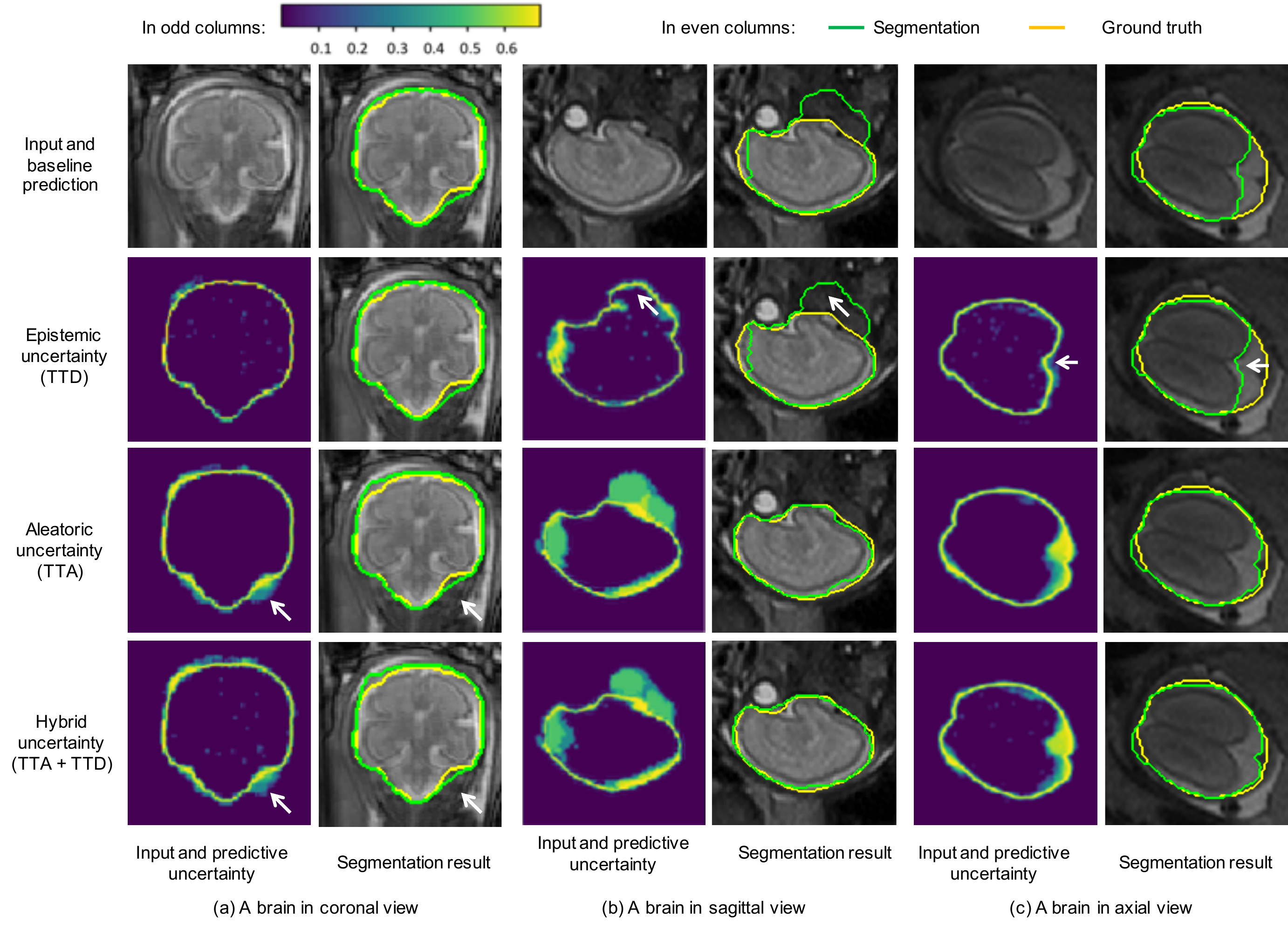}
	\caption{Visual comparison of different types of uncertainties and their corresponding segmentations for fetal brain. The uncertainty maps in odd columns are based on Monte Carlo simulation with $N=20$ and encoded by the color bar in the left up corner (low uncertainty shown in purple and high uncertainty shown in yellow). The white arrows in (a) show the \textit{aleatoric} and hybrid uncertainties in a challenging area, and the white arrows in (b) and (c) show mis-segmented regions with very low \textit{epistemic} uncertainty. TTD: test-time dropout, TTA: test-time augmentation.}
	\label{fig:fetal_visual}
\end{figure*}

\subsubsection{Segmentation Results with Uncertainty}
Fig.~\ref{fig:fetal_visual} shows a visual comparison of different types of uncertainties for segmentation of three fetal brain images in coronal, sagittal and axial view respectively. 
The results were based on the same trained model of U-Net with train-time augmentation, and the Monte Carlo simulation number $N$ was 20 for TTD, TTA, and TTA + TTD to obtain \textit{epistemic}, \textit{aleatoric} and hybrid uncertainties respectively. 
In each subfigure, the first row presents the input and the segmentation obtained by the single-prediction baseline. The other rows show these three types of uncertainties and their corresponding segmentation results respectively. The uncertainty maps in odd columns are represented by pixel-wise entropy of $N$ predictions and encoded by the color bar in the left top corner. In the uncertainty maps, purple pixels have low uncertainty values and yellow pixels have high uncertainty values.
Fig.~\ref{fig:fetal_visual}(a) shows a fetal brain in coronal view. In this case, the baseline prediction method achieved a good segmentation result. It can be observed that for \textit{epistemic} uncertainty calculated by TTD,  most of the uncertain segmentations are located near the border of the segmented foreground, while the pixels with a larger distance to the border have a very high confidence (i.e., low uncertainty). In addition, the \textit{epistemic} uncertainty map contains some random noise in the brain region. In contrast, the \textit{aleatoric} uncertainty obtained by TTA contains less random noise and it shows uncertain segmentations not only on the border but also in some challenging areas in the lower right corner, as highlighted by the white arrows. In that region, the result obtained by TTA has an over-segmentation, and this is corresponding to the high values in the same region of the \textit{aleatoric} uncertainty map.
The hybrid uncertainty calculated by TTA + TTD is a mixture of \textit{epistemic} and \textit{aleatoric} uncertainty. As shown in the last row of Fig.~\ref{fig:fetal_visual}(a), it looks similar to the \textit{aleatoric} uncertainty map except for some random noise. 

Fig.~\ref{fig:fetal_visual}(b) and Fig.~\ref{fig:fetal_visual}(c) show two other cases where the single-prediction baseline obtained an over-segmentation and an under-segmentation respectively. It can be observed that the \textit{epistemic} uncertainty map shows a high confidence (low uncertainty) in these mis-segmented regions. This leads to a lot of overconfident incorrect segmentations, as highlighted by the white arrows in Fig.~\ref{fig:fetal_visual}(b) and (c). In comparison, the \textit{aleatoric} uncertainty map obtained by TTA shows a larger uncertain area that is mainly corresponding to mis-segmented regions of the baseline. In these two cases, The hybrid uncertainty also looks similar to the \textit{aleatoric} uncertainty map. The comparison indicates that the \textit{aleatoric} uncertainty has  a better ability than the \textit{epistemic} uncertainty to indicate mis-segmentations of non-border pixels. For these pixels, the segmentation output is more affected by different transformations of the input (\textit{aleatoric}) rather than variations of model parameters (\textit{epistemic}).   

Fig.~\ref{fig:fetal_visual}(b) and (c) also show that TTD using different model parameters seemed to obtain very little improvement from the baseline. In comparison, TTA using different input transformations corrected the large mis-segmentations and achieved a more noticeable improvement from the baseline. It can also be observed that the results obtained by TTA + TTD are very similar to those obtained by TTA, which shows TTA is more suitable to improving the segmentation than TTD. 
\subsubsection{Quantitative Evaluation}
\begin{figure*}[t]
	\centering
	\includegraphics[width=1.0\linewidth]{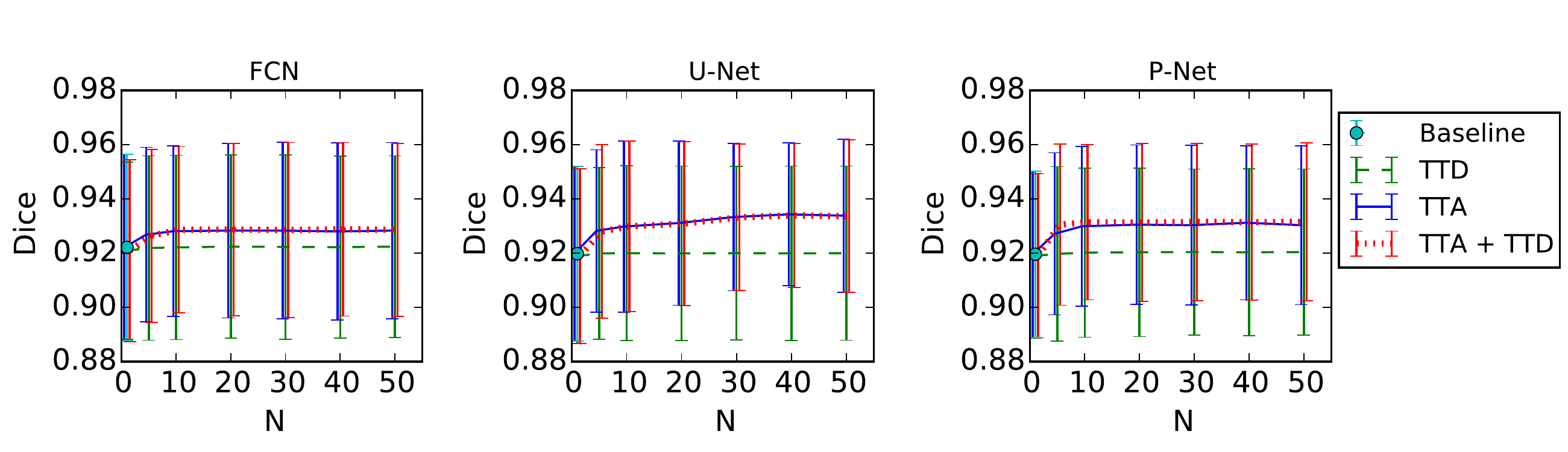}
	\caption{Dice of 2D fetal brain segmentation with different $N$ that is the number of Monte Carlo simulation runs.}
	\label{fig:dice_with_n}
\end{figure*}
\begin{figure*}[t]
	\centering
	\includegraphics[width=0.9\linewidth]{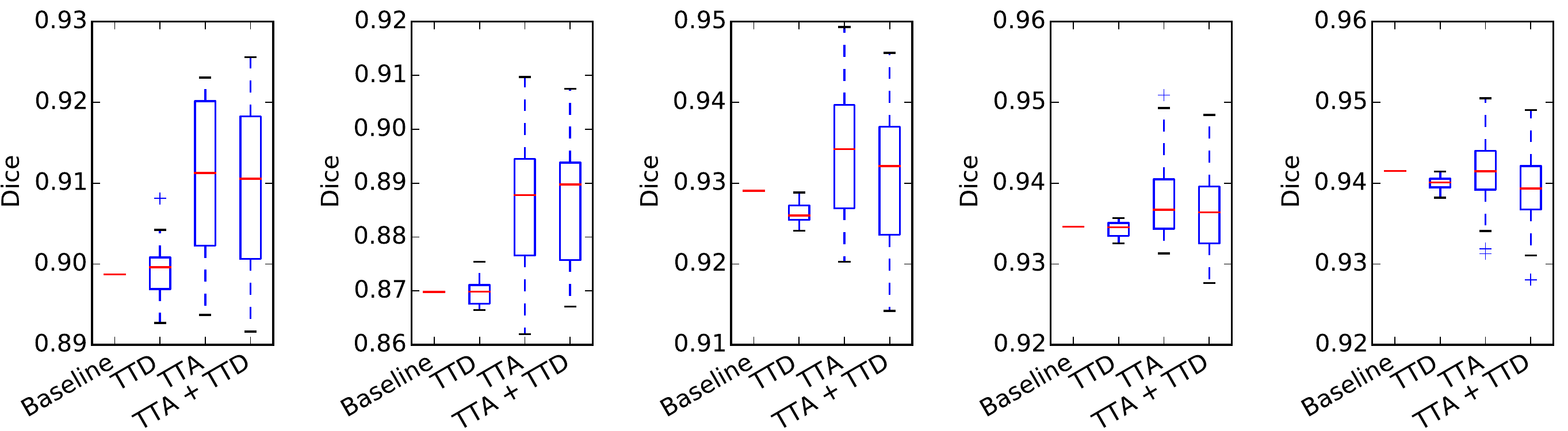}
	\caption{Dice distributions of segmentation results with different testing methods for five example stacks of 2D slices of fetal brain MRI. Note TTA's higher mean value and variance compared with TTD.}
	\label{fig:fetal_5_case}
\end{figure*}

\begin{table*}[t]
	\caption{Dice (\%) and ASSD (mm) evaluation of 2D fetal brain segmentation with different training and testing methods. Tr-Aug: Training without data augmentation. Tr+Aug: Training with data augmentation. * denotes significant improvement from the baseline of single prediction in Tr-Aug and Tr+Aug respectively ($p$-value $<$ 0.05). $\dagger$ denotes significant improvement from Tr-Aug with TTA + TTD ($p$-value $<$ 0.05).}
	\label{tab:2d_dice_assd}
	\centering
	\small
	\begin{tabular}{llllllll}
		\toprule
		\multirow{2}{*}{Train} & \multirow{2}{*}{Test} & \multicolumn{3}{c}{Dice (\%)}   &     \multicolumn{3}{c}{ASSD (mm)}           \\
		\cmidrule(lr){3-5} \cmidrule(lr){6-8}
		 &  & FCN & U-Net & P-Net &  FCN & U-Net & P-Net \\
		\midrule
		
		\multirow{4}{*}{Tr-Aug} & Baseline & 91.05$\pm$3.82 & 90.26$\pm$4.77 &90.65$\pm$4.29 & 2.68$\pm$2.93 & 3.11$\pm$3.34 & 2.83$\pm$3.07   \\
		&TTD  & 91.13$\pm$3.60 & 90.38$\pm$4.30 &90.93$\pm$4.04 & 2.61$\pm$2.85 & 3.04$\pm$2.29 & 2.69$\pm$2.90   \\	
		&TTA & 91.99$\pm$3.48* & 91.64$\pm$4.11* &92.02$\pm$3.85* & 2.26$\pm$2.56* & 2.51$\pm$3.23* & 2.28$\pm$2.61*   \\
		&TTA + TTD & \bf{92.05$\pm$3.58}* & \bf{91.88$\pm$3.61}* & \bf{92.17$\pm$3.68}* & \bf{2.19$\pm$2.67}* & \bf{2.40$\pm$2.71}* & \bf{2.13$\pm$2.42}*  \\ 
		\midrule
		\multirow{4}{*}{Tr+Aug}&	Baseline & 92.03$\pm$3.44 & 91.93$\pm$3.21 &91.98$\pm$3.92 & 2.21$\pm$2.52 & 2.12$\pm$2.23 & 2.32$\pm$2.71   \\
		&TTD  & 92.08$\pm$3.41 & 92.00$\pm$3.22 &92.01$\pm$3.89 & 2.17$\pm$2.52 & 2.03$\pm$2.13 & 2.15$\pm$2.58   \\	
		&TTA & 92.79$\pm$3.34* & 92.88$\pm$3.15* &93.05$\pm$2.96* & 1.88$\pm$2.08 & 1.70$\pm$1.75 & 1.62$\pm$1.77*   \\
		& TTA + TTD & \bf{92.85$\pm$3.15}*$\dagger$ & \bf{92.90$\pm$3.16}*$\dagger$ & \bf{93.14$\pm$2.93}*$\dagger$ & \bf{1.84$\pm$1.92} & \bf{1.67$\pm$1.76}*$\dagger$ & \bf{1.48$\pm$1.63}*$\dagger$  \\ 
		\bottomrule
	\end{tabular}
\end{table*}
To quantitatively evaluate the segmentation results, we measured Dice score and ASSD of predictions by different testing methods with three network structures: FCN~\citep{Long2014}, U-Net~\citep{Hefny2015a} and P-Net~\citep{Wang2018}. For all of these CNNs, we used data augmentation at training time to enlarge the training set. At inference time, we compared the baseline testing method (without Monte Carlo simulation) with TTD, TTA and TTA + TTD. We first investigated how the segmentation accuracy changes with the increase of the number of Monte Carlo simulation runs $N$. The results measured with all the testing images are shown in Fig.~\ref{fig:dice_with_n}. 
We found that for all of these three networks, the segmentation accuracy of TTD remains close to that of the single-prediction baseline. For TTA and TTA + TTD, an improvement of segmentation accuracy can be observed when $N$ increases from 1 to 10. When $N$ is larger than 20, the segmentation accuracy for these two methods reaches a plateau.

In addition to the previous scenario using augmentation at both training and test time,
we also evaluated the performance of TTD and TTA when data augmentation was not used for training. The quantitative evaluations of combinations of different training methods and testing methods ($N = $20) are shown in Table~\ref{tab:2d_dice_assd}.  
It can be observed that for both training with and without data augmentation, TTA has a better ability to improve the segmentation accuracy than TTD. Combining TTA and TTD can further improve the segmentation accuracy, but it does not significantly outperform TTA ($p$-value $>$ 0.05). 

Fig.~\ref{fig:fetal_5_case} shows Dice distributions of five example stacks of fetal brain MRI. The results were based on the same trained model of U-Net with train-time augmentation. Note that the baseline had only one prediction for each image, and the Monte Carlo simulation number $N$ was 20 for TTD, TTA and TTA + TTD. It can be observed that for each case, the Dice of TTD is distributed closely around that of the baseline. In comparison, the Dice distribution of TTA has a higher average than that of TTD, indicating TTA's better ability of improving segmentation accuracy. The results of TTA also have a larger variance than that of TTD, which shows TTA can provide more structure-wise uncertainty information. Fig.~\ref{fig:fetal_5_case} also shows that the performance of TTA + TTD is close to that of TTA.

\subsubsection{Correlation between Uncertainty and Segmentation Error}

\begin{figure*}[t]
	\centering
	\includegraphics[width=1.0\linewidth]{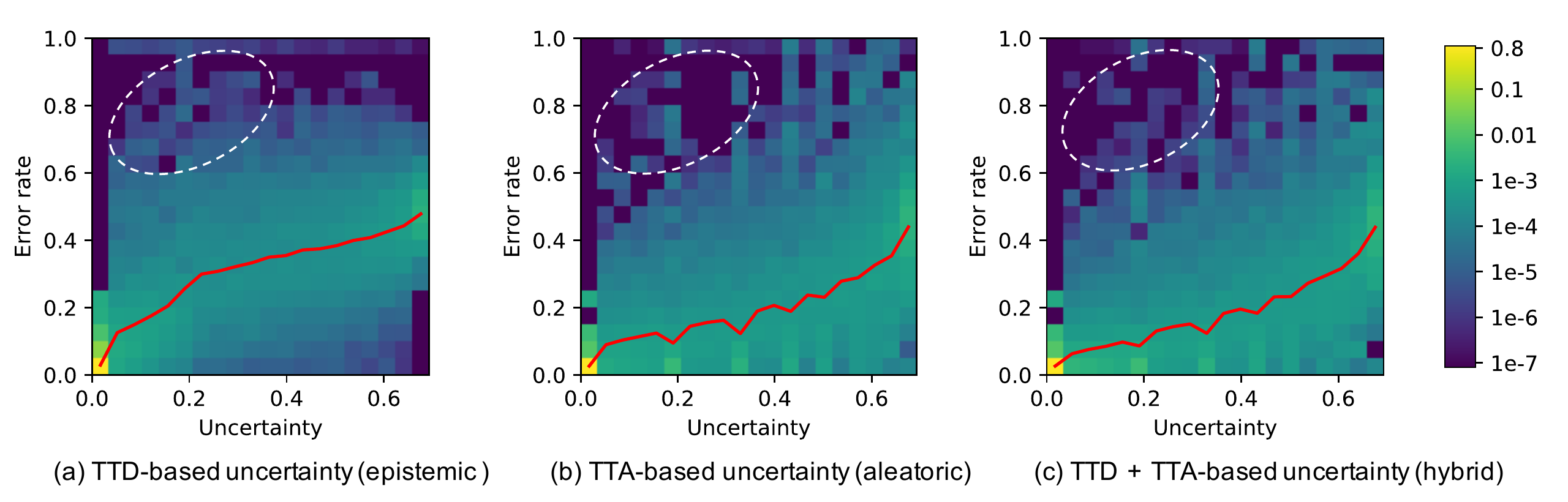}
	\caption{Normalized joint histogram of prediction uncertainty and error rate for 2D fetal brain segmentation. The average error rates at different uncertainty levels are depicted by the red curves. The dashed ellipses show that TTA leads to a lower occurrence of overconfident incorrect predictions than TTD. }
	\label{fig:error_uncertain}
\end{figure*}

\begin{figure*}
	\centering
	\begin{subfigure}[1]{0.28\linewidth}
		\includegraphics[width=\linewidth]{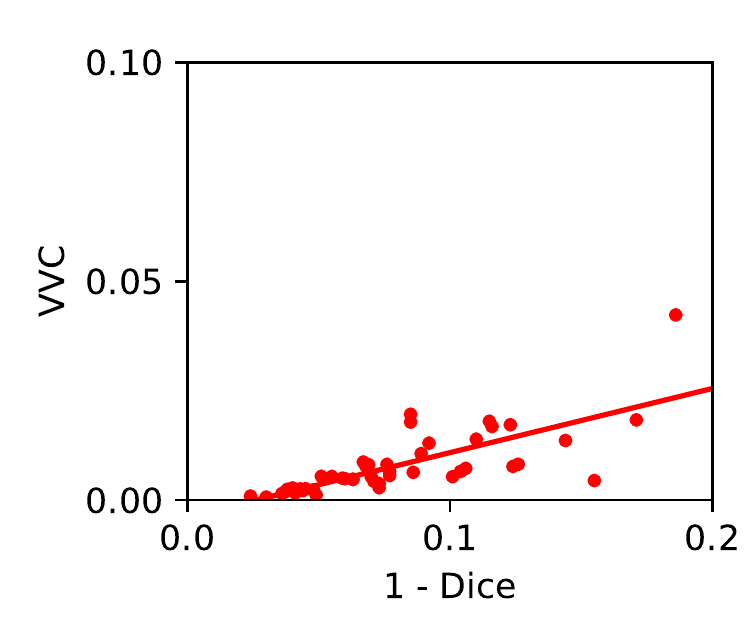}
		\caption{TTD}
		\label{fig:fetal_dice_vvc_ttd}
	\end{subfigure}
~
	\begin{subfigure}[2]{0.28\linewidth}
		\includegraphics[width=\linewidth]{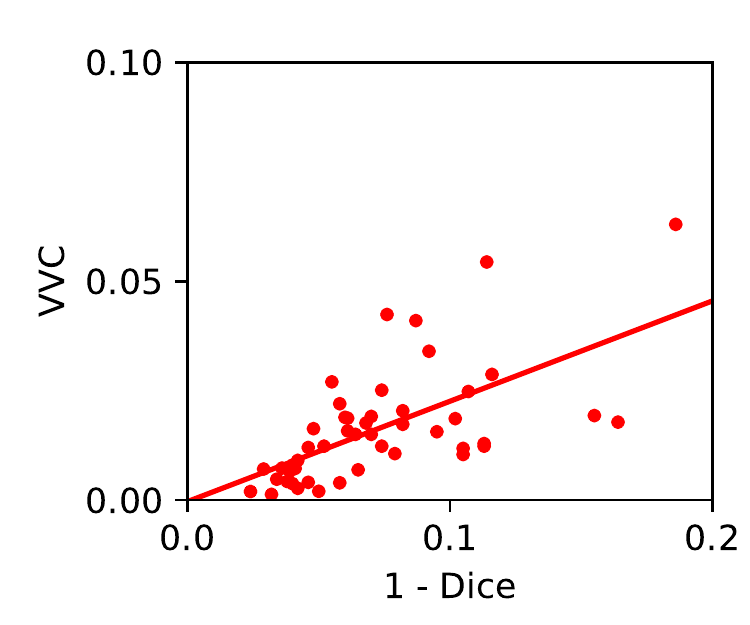}
		\caption{TTA}
		\label{fig:fetal_dice_vvc_tta}
	\end{subfigure}
~
	\begin{subfigure}[2]{0.28\linewidth}
		\includegraphics[width=\linewidth]{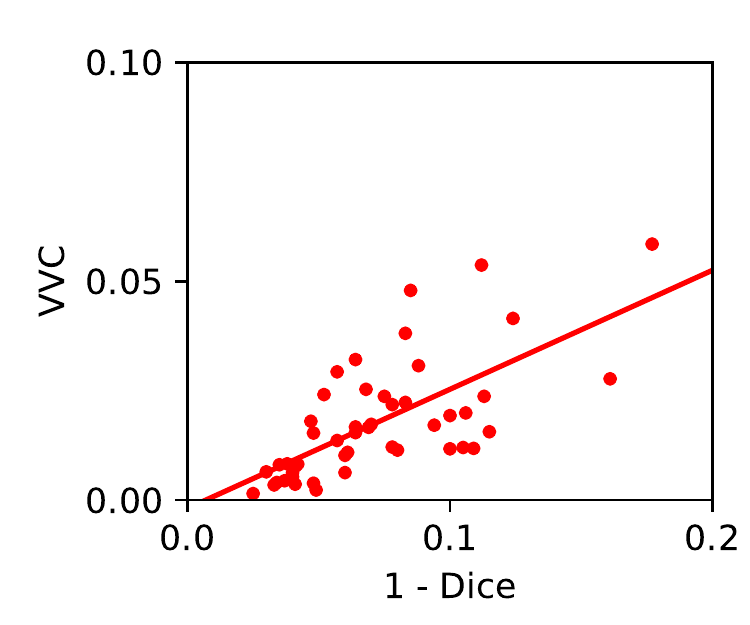}
		\caption{TTA + TTD}
		\label{fig:fetal_dice_vvc_ttad}
	\end{subfigure}
	\caption{Structure-wise uncertainty in terms of volume variation coefficient (VVC) vs $1 - $Dice for different testing methods in 2D fetal brain segmentation. }\label{fig:fetal_s_wise_uncertainty}
\end{figure*}
To investigate how our uncertainty estimation methods can indicate incorrect segmentation, we measured the uncertainty and segmentation error at both pixel-level and structure-level. For pixel-level evaluation, we measured the joint histogram of pixel-wise uncertainty and error rate for TTD, TTA, and TTA + TTD respectively. The histogram was obtained by statistically calculating the error rate of pixels at different pixel-wise uncertainty levels in each slice. The results based on U-Net with $N=20$ are shown in Fig.~\ref{fig:error_uncertain}, where the joint histograms have been normalized by the number of total pixels in the testing images for visualization. For each type of pixel-wise uncertainty, we calculated the average error rate at each pixel-wise uncertainty level, leading to a curve of error rate as a function of pixel-wise uncertainty, i.e., the red curves in Fig.~\ref{fig:error_uncertain}. This figure shows that the majority of pixels have a low uncertainty with a small error rate. When the uncertainty increases, the error rate also becomes higher gradually. Fig.~\ref{fig:error_uncertain}(a) shows the TTD-based uncertainty (\textit{epistemic}). It can be observed that when the prediction uncertainty is low, the result has a steep increase of error rate. In contrast, for the TTA-based uncertainty (\textit{aleatoric}), the increase of error rate is slower, shown in Fig.~\ref{fig:error_uncertain}(b). This demonstrates that TTA has fewer overconfident incorrect predictions than TTD. The dashed ellipses in Fig.~\ref{fig:error_uncertain} also show the different levels of overconfident incorrect predictions for different testing methods. 

For structure-wise evaluation, we used VVC to represent structure-wise uncertainty and $1 - $Dice to represent structure-wise segmentation error. Fig.~\ref{fig:fetal_s_wise_uncertainty} shows the joint distribution of VVC and $1 - $Dice for different testing methods using U-Net trained with data augmentation and $N$ = 20 for inference. The results of TTD, TTA, and TTA + TTD are shown in Fig.~\ref{fig:fetal_s_wise_uncertainty}(a), (b) and (c) respectively. It can be observed that for all the three testing methods, the VVC value tends to become larger when $1 - $Dice grows. However, the slope in Fig.~\ref{fig:fetal_s_wise_uncertainty}(a) is smaller than those in Fig.~\ref{fig:fetal_s_wise_uncertainty}(b) and Fig.~\ref{fig:fetal_s_wise_uncertainty}(c). The comparison shows that TTA-based structure-wise uncertainty estimation is highly related to segmentation error, and TTA leads to a larger scale of VVC than TTD. Combining TTA and TTD leads to similar results to that of TTA.

\subsection{3D Brain Tumor Segmentation from Multi-Modal MRI}\label{3D_exepriment}
MRI has become the most commonly used imaging methods for brain tumors. Different MR sequences such as T1-weighted (T1w), contrast enhanced
T1-weighted (T1wce), T2-weighted (T2w) and Fluid Attenuation Inversion Recovery (FLAIR) images can provide complementary information for analyzing multiple subregions of brain tumors. Automatic brain tumor segmentation from multi-modal MRI has a potential for better diagnosis, surgical planning and treatment assessment~\citep{Menze2015}. Deep neural networks have achieved the state-of-the-art performance on this task~\citep{Kamnitsas2017,Wang17brats}. In this experiment, we analyze the uncertainty of deep CNN-based brain tumor segmentation and show the effect of our proposed test-time augmentation. 
\subsubsection{Data and Implementation}
We used the BraTS 2017\footnote{\url{http://www.med.upenn.edu/sbia/brats2017.html}}~\citep{Bakas2017} training dataset that consisted of volumetric images from 285 studies, with ground truth provided by the organizers. We randomly selected 20 studies for validation and 50 studies for testing, and used the remaining for training. For each study, there were four scans of T1w, T1wce, T2w and FLAIR images, and they had been co-registered. All the
images were skull-stripped and re-sampled to an isotropic 1 mm$^3$ resolution. As a first demonstration of uncertainty estimation for deep learning-based brain tumor segmentation, we investigate segmentation of the whole tumor from these multi-modal images (Fig.~\ref{fig:brain_tumor_visual}). We used 3D U-Net~\citep{Abdulkadir2016}, V-Net~\citep{Milletari2016} and W-Net~\citep{Wang17brats} implemented with NiftyNet~\citep{Gibson2018}, and employed Adam during training with initial learning rate $10^{-3}$, batch size 2, weight decay $10^{-7}$ and iteration number $20k$. W-Net is a 2.5D network, and we compared using W-Net only in axial view and a fusion of axial, sagittal and coronal views. These two implementations are referred to as W-Net(A) and W-Net(ASC) respectively.  The transformation parameter $\bm\beta$ in the proposed augmentation framework consisted of $f_l$, $r$, $s$ and $e$, where $f_l$ is a random variable for flipping along each 3D axis, $r$ is the rotation angle along each 3D axis, $s$ is a scaling factor and $e$ is intensity noise. The prior distributions were: $f_l \sim Bern(0.5)$, $r\sim U(0, 2\pi)$, $s\sim U(0.8, 1.2)$ and $ e \sim N(0, 0.05)$ according to the reduced standard deviation of a median-filtered version of a normalized image. We used this formulated augmentation during training, and also employed it to obtain TTA-based results at test time.

\subsubsection{Segmentation Results with Uncertainty}
\begin{figure*}[t]
	\centering
	\includegraphics[width=1.0\linewidth]{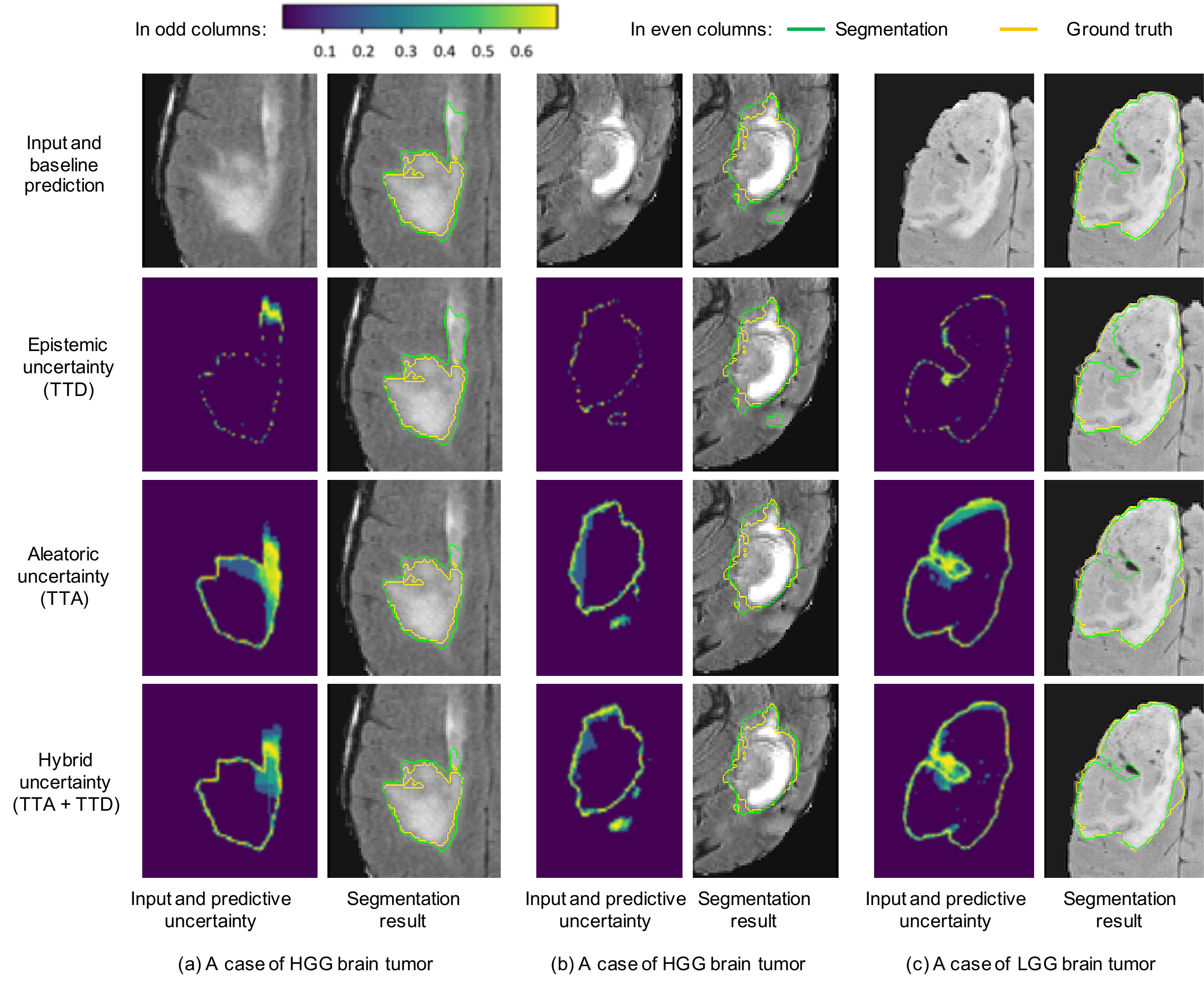}
	\caption{Visual comparison of different testing methods for 3D brain tumor segmentation. The uncertainty maps in odd columns are based on Monte Carlo simulation with $N=40$ and encoded by the color bar in the left up corner (low uncertainty shown in purple and high uncertainty shown in yellow). TTD: test-time dropout, TTA: test-time augmentation.
	}
	\label{fig:brain_tumor_visual}
\end{figure*}
Fig.~\ref{fig:brain_tumor_visual} demonstrates three examples of uncertainty estimation of brain tumor segmentation by different testing methods. The results were based on the same trained model of 3D U-Net~\citep{Abdulkadir2016}. The Monte Carlo simulation number $N$ was 40 for TTD, TTA, and TTA + TTD to obtain \textit{epistemic}, \textit{aleatoric} and hybrid uncertainties respectively. Fig.~\ref{fig:brain_tumor_visual}(a) shows a case of high grade glioma (HGG). The baseline of single prediction obtained an over-segmentation at the upper part of the image. The \textit{epistemic} uncertainty obtained by TTD highlights some uncertain predictions at the border of the segmentation and a small part of the over-segmented region. In contrast, the \textit{aleatoric} uncertainty obtained by TTA better highlights the whole over-segmented region, and the hybrid uncertainty map obtained by TTA + TTD is similar to the  \textit{aleatoric} uncertainty map. The second column of Fig.~\ref{fig:brain_tumor_visual}(a) shows the corresponding segmentations of these uncertainties. It can be observed that the TTD-based result looks similar to the baseline, while TTA and TTA + TTD based results achieve a larger improvement from the baseline. Fig.~\ref{fig:brain_tumor_visual}(b) demonstrates another case of HGG brain tumor, and it shows that the over-segmented region in the baseline prediction is better highlighted by TTA-based \textit{aleatoric} uncertainty than TTD-based \textit{epistemic} uncertainty.  Fig.~\ref{fig:brain_tumor_visual}(c) shows a case of low grade glioma (LGG). The baseline of single prediction  obtained an under-segmentation in the middle part of the tumor. The \textit{epistemic} uncertainty obtained by TTD only highlights pixels on the border of the prediction, with a low uncertainty (high confidence) for the under-segmented region. In contrast, the \textit{aleatoric} uncertainty obtained by TTA has a better ability to indicate the under-segmentation. The results also show that TTA outperforms TTD for better segmentation.

\subsubsection{Quantitative Evaluation}
For quantitative evaluations, we calculated the Dice score and ASSDe for the segmentation results obtained by the different testing methods that were combined with 3D U-Net~\citep{Abdulkadir2016}, V-Net~\citep{Milletari2016} and W-Net~\citep{Wang17brats} respectively. We also compared TTD and TTA with and without train-time data augmentation, respectively.
We found that for these networks, the performance of the multi-prediction testing methods  reaches a plateau when $N$ is larger than 40. Table~\ref{tab:3d_dice_assd} shows the evaluation results with $N$ = 40.    
It can be observed that for each network and each training method, multi-prediction methods lead to better performance than the baseline with a single prediction, and TTA outperforms TTD with higher Dice scores and lower ASSD values. Combining TTA and TTD has a slight improvement from using TTA, but the improvement is not significant ($p$-value $<$ 0.05).

\begin{table*}[t]
	\caption{Dice (\%) and ASSD (mm) evaluation of 3D brain tumor segmentation with different training and testing methods. Tr-Aug: Training without data augmentation. Tr+Aug: Training with data augmentation. W-Net is a 2.5D network and W-Net (ASC) denotes the fusion of axial, sagittal and coronal views according to~\cite{Wang17brats}. * denotes significant improvement from the baseline of single prediction in Tr-Aug and Tr+Aug respectively ($p$-value $<$ 0.05). $\dagger$ denotes significant improvement from Tr-Aug with TTA + TTD ($p$-value $<$ 0.05). }
	\label{tab:3d_dice_assd}
	\centering
	\small
	\begin{tabular}{llllllll}
		\toprule
		\multirow{2}{*}{Train}&\multirow{2}{*}{Test} & \multicolumn{3}{c}{Dice (\%)}   &     \multicolumn{3}{c}{ASSD (mm)}           \\
		\cmidrule(lr){3-5} \cmidrule(lr){6-8}
		&  & WNet (ASC) & 3D U-Net & V-Net &  WNet (ASC) & 3D U-Net & V-Net \\
		\midrule
		\multirow{4}{*}{Tr-Aug}&  Baseline  & 87.81$\pm$7.27 &   87.26$\pm$7.73 &86.84$\pm$8.38 & 2.04$\pm$1.27 & 2.62$\pm$1.48 & 2.86$\pm$1.79   \\
		& TTD   & 88.14$\pm$7.02  & 87.55$\pm$7.33 &87.13$\pm$8.14   & 1.95$\pm$1.20 & 2.55$\pm$1.41 & 2.82$\pm$1.75   \\	
		& TTA & 89.16$\pm$6.48* & 88.58$\pm$6.50* &87.86$\pm$6.97*  &1.42$\pm$0.93* & 1.79$\pm$1.16* & 1.97$\pm$1.40*  \\
		& TTA + TTD & \bf{89.43$\pm$6.14}* & \bf{88.75$\pm$6.34}* & \bf{88.03$\pm$6.56}* &  \bf{1.37$\pm$0.89}* & \bf{1.72$\pm$1.23}* & \bf{1.95$\pm$1.31}*  \\ 
		\midrule
		\multirow{4}{*}{Tr+Aug}&  Baseline  & 88.76$\pm$5.76 &   88.43$\pm$6.67 &87.44$\pm$7.84 & 1.61$\pm$1.12 & 1.82$\pm$1.17 & 2.07$\pm$1.46   \\
		& TTD   & 88.92$\pm$5.73  & 88.52$\pm$6.66 &87.56$\pm$7.78   & 1.57$\pm$1.06 & 1.76$\pm$1.14 & 1.99$\pm$1.33   \\	
		& TTA & 90.07$\pm$5.69* & 89.41$\pm$6.05* &88.38$\pm$6.74*  &1.13$\pm$0.54* & 1.45$\pm$0.81 & 1.67$\pm$0.98*  \\
		& TTA + TTD & \bf{90.35$\pm$5.64}*$\dagger$ & \bf{89.60$\pm$5.95}*$\dagger$ & \bf{88.57$\pm$6.32}*$\dagger$ &  \bf{1.10$\pm$0.49}* & \bf{1.39$\pm$0.76}*$\dagger$ & \bf{1.62$\pm$0.95}*$\dagger$  \\
		\bottomrule
	\end{tabular}
\end{table*} 

\subsubsection{Correlation between Uncertainty and Segmentation Error}
\begin{figure}[t]
	\centering
	\includegraphics[width=1.0\linewidth]{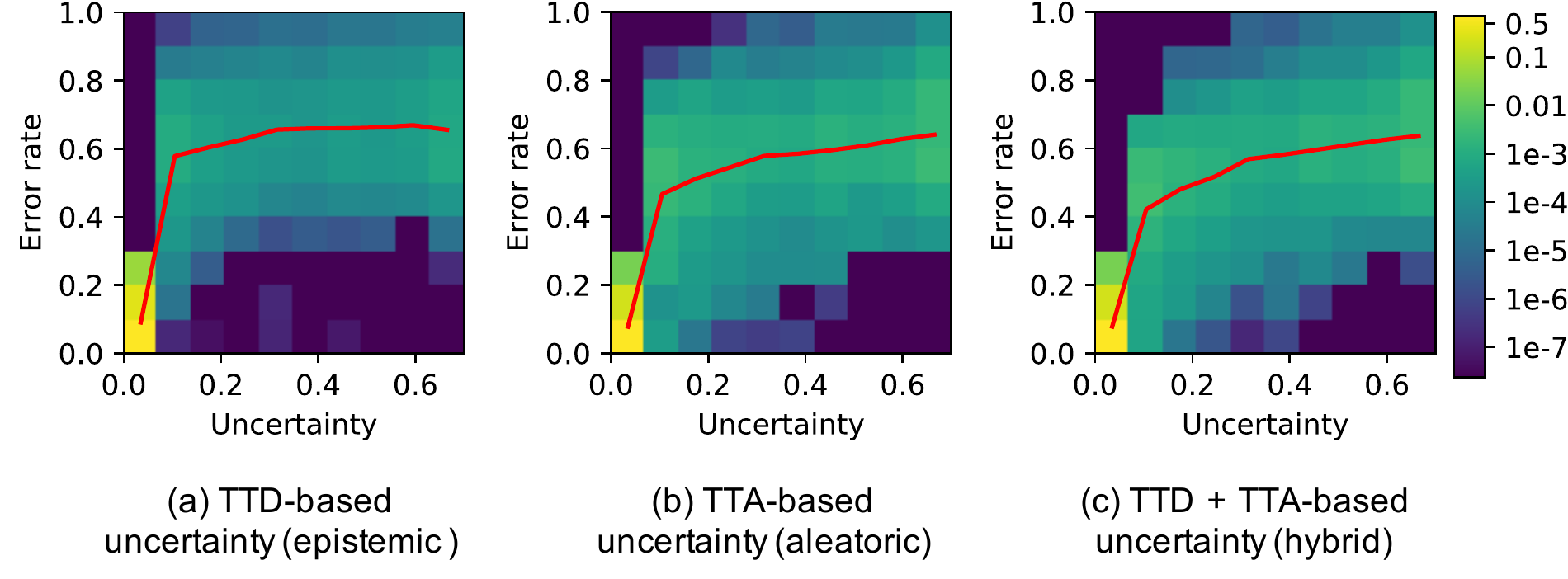}
	\caption{Normalized joint histogram of prediction uncertainty and error rate for 3D brain tumor segmentation. The average error rates at different uncertainty levels are depicted by the red curves.}
	\label{fig:ucertain_error_3d}
\end{figure}

To study the relationship between prediction uncertainty and segmentation error at voxel-level, we measured voxel-wise uncertainty and voxel-wise error rate at different uncertainty levels. For each of TTD-based (\textit{epistemic}),  TTA-based (\textit{aleatoric}) and TTA + TTD-based (hybrid) voxel-wise uncertainty, we obtained the normalized joint histogram of voxel-wise uncertainty and voxel-wise error rate. Fig.~\ref{fig:ucertain_error_3d} shows the results based on 3D U-Net trained with data augmentation and using $N=40$ for inference. The red curve shows the average voxel-wise error rate as a function of voxel-wise uncertainty. In Fig.~\ref{fig:ucertain_error_3d}(a), the average prediction error rate has a slight change when the TTD-based \textit{epistemic} uncertainty is larger than 0.2. In contrast, Fig.~\ref{fig:ucertain_error_3d}(b) and (c) show that the average prediction error rate has a smoother increase with the growth of \textit{aleatoric} and hybrid uncertainties. The comparison demonstrates that the TTA-based \textit{aleatoric} uncertainty leads to fewer over-confident mis-segmentations than the TTD-based \textit{epistemic} uncertainty. 

\begin{figure}
	\centering
	\begin{subfigure}[1]{0.31\linewidth}
		\includegraphics[width=\linewidth]{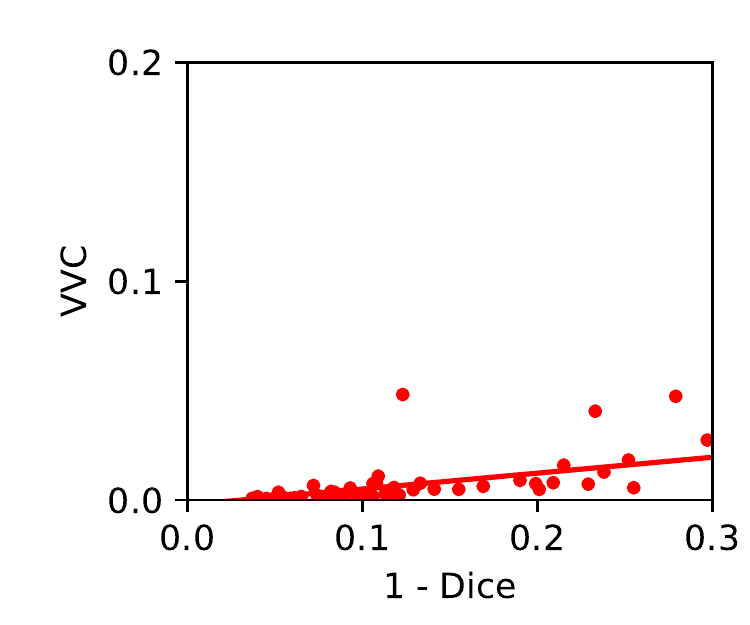}
		\caption{TTD}
		\label{fig:glioma_dice_vvc_ttd}
	\end{subfigure}
	~
	\begin{subfigure}[2]{0.31\linewidth}
		\includegraphics[width=\linewidth]{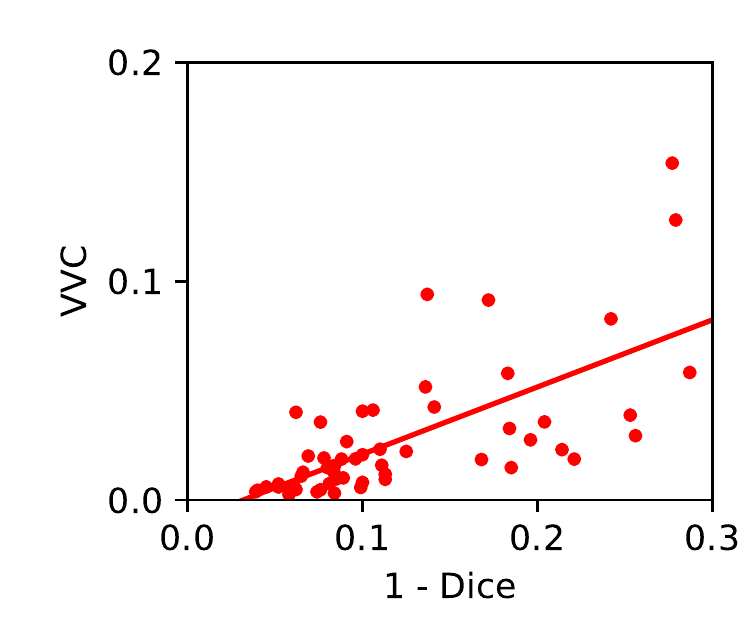}
		\caption{TTA}
		\label{fig:glioma_dice_vvc_tta}
	\end{subfigure}
	~
	\begin{subfigure}[2]{0.31\linewidth}
		\includegraphics[width=\linewidth]{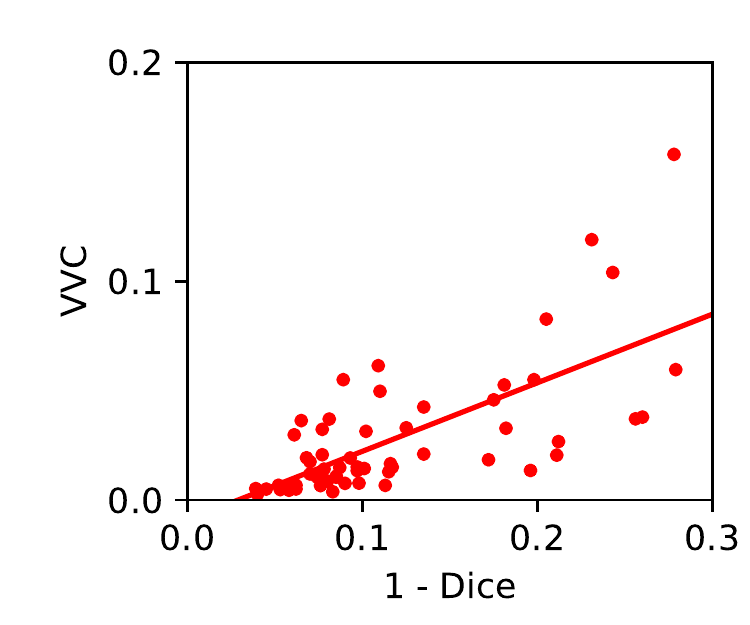}
		\caption{TTA + TTD}
		\label{fig:glioma_dice_vvc_ttad}
	\end{subfigure}
	\caption{Structure-wise uncertainty in terms of volume variation coefficient (VVC) vs $1 - $Dice for different testing methods in 3D brain tumor segmentation. }\label{fig:glioma_dice_vvc}
\end{figure}

For structure-level evaluation, we also studied the relationship between structure-level uncertainty represented by VVC and structure-level error represented by $1 - $Dice. Fig.~\ref{fig:glioma_dice_vvc} shows their joint distributions with three different testing methods using 3D U-Net. The network was trained with data augmentation, and $N$ was set as 40 for inference. Fig.~\ref{fig:glioma_dice_vvc} shows that TTA-based VVC increases when $1 - $ Dice grows, and the slope is larger than that of TTD-based VVC. The results of TTA and TTA + TTD are similar, as shown in Fig.~\ref{fig:glioma_dice_vvc}(b) and (c). The comparison shows that TTA-based structure-wise uncertainty can better indicate segmentation error than TTD-based structure-wise uncertainty. 
    
\section{Discussion and Conclusion}\label{sec:discussion}

In our experiments, the number of training images was relatively small compared with many datasets of natural images such as PASCAL VOC, COCO and ImageNet. For medical images, it is typically very difficult to collect a very large dataset for segmentation, as pixel-wise annotations are not only time-consuming to collect but also require expertise of radiologists. Therefore, for most existing medical image segmentation datasets, such as those in Grand challenge\footnote{ https://grand-challenge.org/challenges}, the image numbers are also quite small. Therefore, investigating the segmentation performance of CNNs with limited training data is of high interest for medical image computing community. In addition, our dataset is not very large so that it is suitable for data augmentation, which fits well with our motivation of using data augmentation at training and test time. The need for uncertainty estimation is also stronger in cases where datasets are smaller.

In our mathematical formulation of test-time augmentation based on an image acquisition model, we explicitly modeled spatial transformations and image noise. However, it can be easily extended to include more general transformations such as elastic deformations~\citep{Abdulkadir2016} or add a simulated bias field for MRI. 
In addition to the variation of possible values of model parameters, the prediction result is also dependent on the input data, e.g., image noise and transformations related to the object. Therefore, a good uncertainty estimation should take these factors into consideration.  Fig.~\ref{fig:fetal_visual} and~\ref{fig:brain_tumor_visual} show that model uncertainty alone is likely to obtain overconfident incorrect predictions, and TTA plays an important role in reducing such predictions. 
In Fig.~\ref{fig:fetal_5_case} we show five example cases, where each subfigure shows the results for one patient. Table~\ref{tab:2d_dice_assd} shows the statistical results based on all the testing images. We found that for few testing images TTA +TTD failed to obtain higher Dice scores than TTA, but for the overall testing images, the average Dice of TTA + TTD is slightly larger than that of TTA. Therefore, this leads to the conclusion that TTA + TTD does not always perform better than TTA, and the average performance of TTA + TTD is close to that of TTA, which is also demonstrated in Fig.~\ref{fig:fetal_visual} and~\ref{fig:brain_tumor_visual}.

We have demonstrated TTA based on the image acquisition model for image segmentation tasks, but it is general for different image recognition tasks, such as image classification, object detection, and regression. For regression tasks where the outputs are not discretized category labels, the variation of the output distribution might be more suitable than entropy for uncertainty estimation. 
Table~\ref{tab:3d_dice_assd} shows the superiority of test-time augmentation for better segmentation accuracy, and it also demonstrates the combination of W-Net in different views helps to improve the performance. This is an ensemble of three networks, and such an ensemble may be used as an alternative for \textit{epistemic} uncertainty estimation, as demonstrated by ~\cite{Lakshminarayanan2017}.

We found that for our tested CNNs and applications, the proper value of Monte Carlo sample $N$ that leads the segmentation accuracy to a plateau was around 20 to 40. Using an empirical value $N = 40$ is large enough for our datasets. However, the optimal setting of  hyper-parameter $N$ may change for different datasets. Fixing $N = 40$ for new applications where the optimal value of $N$ is smaller would lead to unnecessary computation and reduce efficiency. In some applications where the object has more spatial variations, the optimal $N$ value may be larger than 40. Therefore, in a new application, we suggest that the optimal $N$ should be determined by the performance plateau on the validation set.   

In conclusion, we analyzed different types of uncertainties for CNN-based medical image segmentation by comparing and combining model (\textit{epistemic}) and input-based (\textit{aleatoric}) uncertainties. We formulated a test-time augmentation-based \textit{aleatoric} uncertainty estimation for medical images that considers the effect of both image noise and spatial transformations. We also proposed a  theoretical and mathematical formulation of test-time augmentation, where we obtain a distribution of the prediction by using Monte Carlo simulation and modeling prior distributions of parameters in an image acquisition model. Experiments with 2D and 3D medical image segmentation tasks showed that uncertainty estimation with our formulated TTA helps to reduce overconfident incorrect predictions encountered by model-based uncertainty estimation and TTA leads to higher segmentation accuracy than a single-prediction baseline and multiple predictions using test-time dropout.

\section{Acknowledgements}\label{sec:acknowledgements}
This work was supported by the Wellcome/EPSRC Centre for Medical Engineering [WT 203148/Z/16/Z], an Innovative Engineering for Health award by the Wellcome Trust (WT101957); Engineering and Physical Sciences Research Council (EPSRC) (NS/A000027/1, EP/H046410/1, EP/J020990/1, EP/K005278), Wellcome/EPSRC [203145Z/16/Z], the National Institute for Health Research University College London Hospitals Biomedical Research Centre (NIHR BRC UCLH/UCL), the Royal Society [RG160569], and hardware donated by NVIDIA.

\bibliographystyle{model2-names}
\bibliography{./reference/midl2018}






\end{document}